# IMAGE SUBSET SELECTION USING GABOR FILTERS AND NEURAL NETWORKS

Heider K. Ali[1] and Anthony Whitehead[2]

[1]Carleton University, Systems & Computer Engineering Department , Ottawa, ON, K1S 5B8, Canada
heider@sce.carleton.ca
[2]Carleton University, School of Information Technology, Ottawa , ON, K1S 5B8, Canada
Anthony.Whitehead@carleton.ca

## ABSTRACT

*An automatic method for the selection of subsets of images, both modern and historic, out of a set of landmark large images collected from the Internet is presented in this paper. This selection depends on the extraction of dominant features using Gabor filtering. Features are selected carefully from a preliminary image set and fed into a neural network as a training data. The method collects a large set of raw landmark images containing modern and historic landmark images and non-landmark images. The method then processes these images to classify them as landmark and non-landmark images. The classification performance highly depends on the number of candidate features of the landmark.*

## 1. INTRODUCTION

Classification of a large set of images containing both modern and historic images allowing a selection of a subset of these images that meets certain technical criteria is based completely on the detection and extraction of image features. Feature extraction has received significant effort of researchers because of its great role in computer vision, image processing, and robotics fields [1]. Many standard feature extraction techniques have been applied to images to achieve optimal extraction performance and to maintain robustness by addressing the varying spatial resolutions, illumination, observer viewpoint, and rotation [2, 3, 4, 5]. However, matching historical images that have been digitized with those taken with modern digital cameras still proves an elusive problem [6].

Some success has been found using a hybrid of standard techniques [6] by merging the ORB detector with the SURF descriptor for matching modern to historic images. This work offered a thorough study of the problems in capturing historic images and the different image captures conditions between modern to historic images. Limited success was achieved in fulfilling the goal of modern to historic image matching using this technique because of the ORB/SURF hybrid technique partially addressed the matching process since the objects of interest in both modern and historic images have to be defined and extracted accurately in terms of pixel values and area levels.

These techniques have achieved partial success in the feature matching in a mixed set of modern and historic images.

Object recognition techniques based on Gabor filters for feature extraction have shown moderate success in extracting fundamental frequencies which represents the shape of an object [7]. Because the Gabor filters act as edge, shape and line detectors, as well as the tuning flexibility of different orientations and frequencies [8], these filters are often applied in a wide range of applications such as texture segmentation and classification [9], face recognition [10], fingerprint matching [11], and motion tracking [12].

In this work, a mixture of modern and historic images captured with cameras of different technologies under different environmental conditions is considered. It is necessary to select the most appropriate image subset from a huge image set with significant redundancy in order to facilitate detailed feature matching in a subsequent step. This task can be successfully achieved by using Gabor filters for feature extraction which offer promising prospects in object recognition where the scale, rotation, translation and illumination invariant recognition can be realized within a reasonable computational time limit [8]. The selected features are trained on a neural network to recognize and extract the true features and to select the images of highest similarity.

This paper presents an automatic matching method to select an image subset of landmarks out from a huge image set downloaded or collected from image websites like Google images or Flickr.

## 2. RELATED WORK

Selection of image subsets from a larger set depends widely on the image type and contents. Thorough analysis of image contents has to be done to extract the salient features and characteristics that are common in an image set. This extraction depends on the conditions under which the image were taken [1] .Traditional image set selection and classification mainly relies on the analysis of the low-level features of the image to determine high-level content semantics [14]. To optimize the feature detection and extraction process, the researchers have worked to support their efforts using non-traditional techniques. Machine learning techniques like artificial neural networks (ANN) has obtained popularity in image subset selection [15].

Xiong et. al. [16] proposed a back propagation neural network to improve the performance of image selection by segmenting and clustering the image into several visual objects and building the total feature vector, while Xu and Qu [17] developed a method based on feature matching similarity and frequency–inverse document frequency (TF-IDF). Multi-scale features used by Li and Zhao[18] in medical image classification where the classifiers use a set of complementary image features in various scales to compare the results of classification process. To optimize the feature selection process for a robust image classification, Al-Sahaf et. al. [19] introduced a Two-Tier Genetic Programming (GP) based image classification method which works on raw pixels rather than high-level features, while Kharrat et. al. [20] based also on the genetic algorithm for the selection of MR brain images using wavelet co-occurrence.

Wrapper feature extraction method has been used by Zhuo et.al. [21] to evaluate of the goodness of selected feature subsets depending on the classification accuracy and they used a Genetic Algorithm (GA) based on the wrapper method for the classification of hyper spectral data using Support Vector Machine(SVM). I-vectors, which are vectors used widely in speaker identification convey the speaker characteristics, adapted by Smith [22] to be used in image classification depended on SURF image features by training a universal Gaussian mixture model (UGMM) on extracted SURF features. To classify the synthetic aperture radar (SAR) images, Feng et. al. [23] developed an approach that takes advantage of both texture and amplitude features in the image based on the superpixels obtained with some over-segmentation methods and applying the support vector machine (SVM). To classify the multispectral images, Long and Singh [24] have introduced a robust algorithm adopting the entropy theory by comparing remotely sensed multispectral images with unknown pixels.

## 3. GABOR FILTERS

Gabor filters are based on the Gabor wavelets which are formed from a complex sinusoidal carrier placed under a Gaussian envelope. These wavelets are based on the Gabor elementary function presented by D. Gabor in 1946 [13]. Many forms of the 2-D Gabor filter have been presented. The 2-D Gabor filter $G(x, y)$ can be defined as [7]:

$$G(x, y) = e^{-(\alpha^2 x_p^2 + \beta^2 y_p^2)} e^{j2\pi f_0 x_p} \tag{1}$$

Where α is the time duration of the Gaussian envelope and the plane wave, $f_0$ is the frequency of the carrier, $x_p = x\cos\theta + y\sin\theta$, $y_p = -x\sin\theta + y\cos\theta$ and Θ and β are the sharpness values of the major and minor axes of the Gaussian envelope. Gabor filters can be used effectively to make the classification with varying capture conditions [8]. In this work we examine the effectiveness of Gabor filters in a situation where other techniques have failed significantly [6]. Specifically, by using an image set of a landmark containing both historic and modern images captured by different technologies widely spread throughout time.

## 4. NEURAL NETWORKS

A multi-layer feed-forward neural network has been used in training a data set on input patterns. A three layer network was used: the input layer, a hidden layer and the output layer. For the input layer, the input feature vectors comes from the Gabor filter feature extraction stage and consists of 100 neurons applied to the neural network. A single pattern *p* to be tested or fed to the neural network can be considered as a vertical vector of elements (features). Then *t* is called the target for this pattern and you have to know *t* in advance and to use it along with its conjugate pattern for the purpose of network training.

For many features in an image set, the set *P* of vertical vectors represents the set of patterns(features) for which you know their desired target in advance in the form of T as horizontal vector. Each element of the *T* vector corresponds to a column in *P* matrix.

These vectors are processed in the hidden layer using the scaled conjugate gradient method as the training method and the mean square error with regularization as the performance function to adjust the network output to be in the range -1.0 to 1.0. The output layer size depends on the obvious candidate features in each landmark. The following figure shows a typical multi-layer feed-forward neural network.

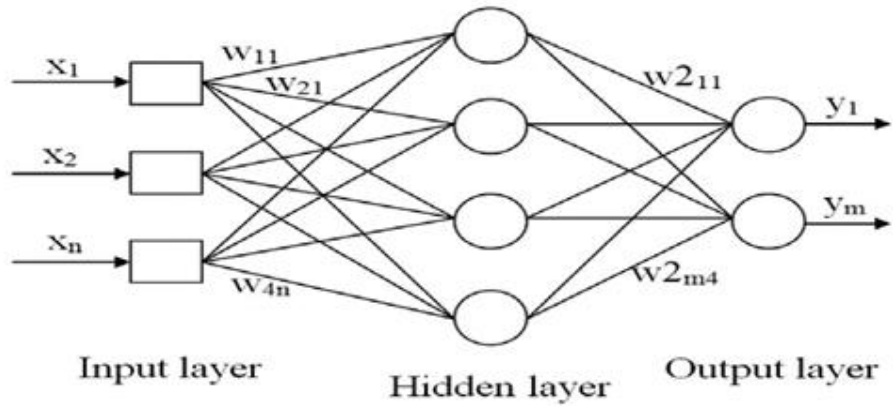

Figure 1. Multi-layer feed-forward neural network

# 5. DATA SUBSET SELECTION

## 5.1. Data sets

The data sets are divided into two categories: the preliminary data sets and the actual dataset. The preliminary data sets are sets of modern and historic images for five landmarks. Every data set contains 300-350 images. This data set is primarily used as a training set to the neural network after the Gabor features has been extracted and fed into the neural network. Figure 2 shows some the images of the preliminary data set for Eiffel tower and Coliseum landmarks. Images of other three are shown in Appendix 1.

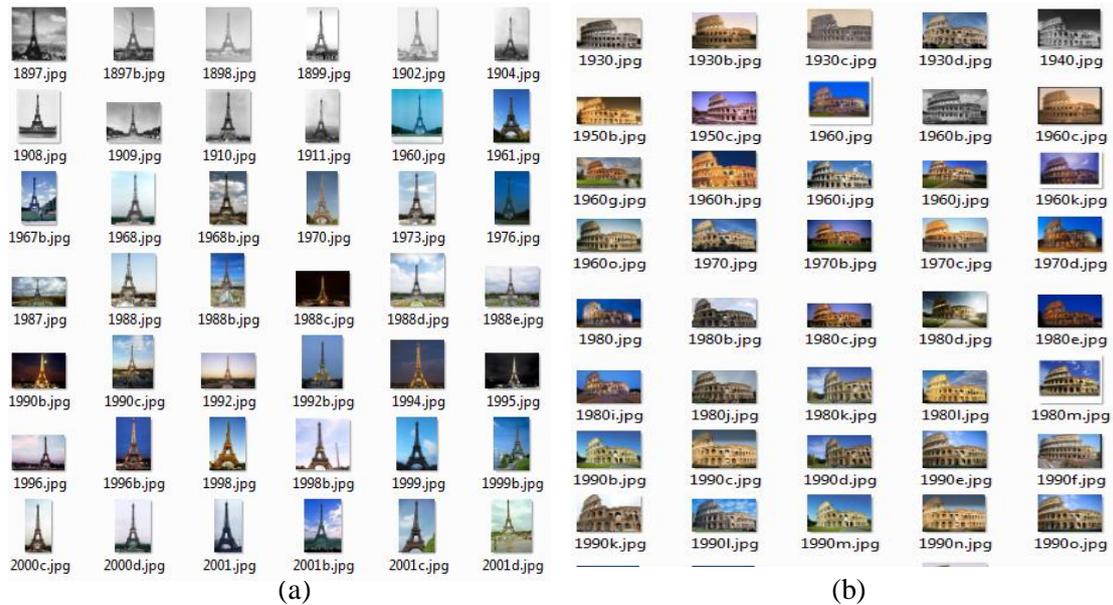

Figure 2. Examples of modern and historic images of many landmarks (a) Eiffel tower in Paris (b) Coliseum of Rome

To prepare these images for the next processing steps, these images have to be converted into gray scale images and resized to a common size. To overcome the problems of variations in brightness and contrast, as well as, the different lighting conditions, histogram equalization [25] was used. Specifically, adaptive histogram equalization (AHE) is used. AHE computes many histograms; each histogram corresponds to a specific sector of the image and uses them to redistributes the image brightness values [26]. Finally, the input values are normalized to a range of [-1, 1].

The actual data set is comprised of sets of modern and historic images of five landmarks. Every set is 1000-3000 images gathered from Google Image Search. These images are placed in a folder, resized, converted to gray scale, and preprocessed to compensate for contrast and lighting problems. These images are applied to the system to classify them as matching or non-matching depending on the Gabor features of each landmark.

## 5.2. Feature extraction by Gabor filtering

To extract the image Gabor filters, it is required to create the Gabor wavelets. These wavelets are created using equation (1) above by using three control factors to create the Gabor kernels. Figure 3 shows the Gabor wavelets generated with three control parameters: scale, frequency and orientation.

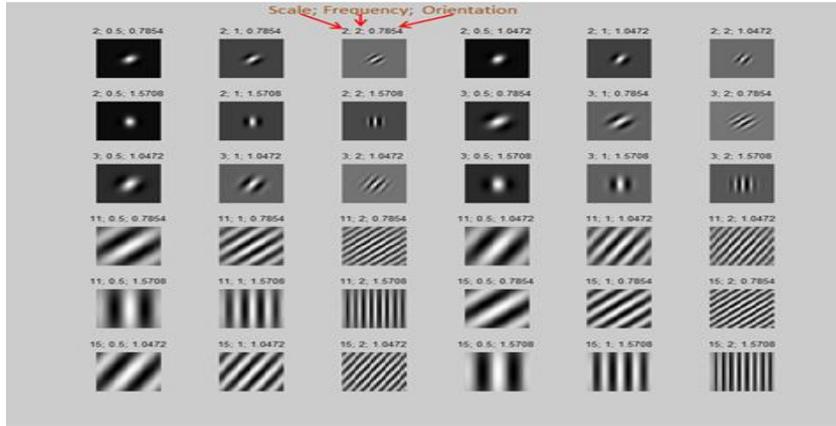
Figure 3. Gabor wavelets created using three control parameters

By applying these kernels on an image, we get the Gabor wavelets family shown in figure 4.

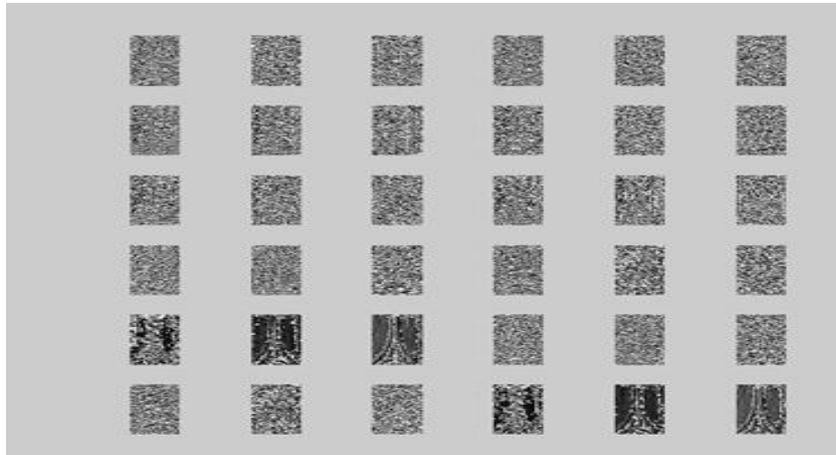
Figure 4. Gabor family

### 5.3. Neural network training

In the training phase of the neural network two sets of images are applied, a feature image set and non-feature image set. The feature image set consists of 100-150 images containing the feature to be tested, while the non-feature image set is 50 non feature images. Feature images completely depend on the landmark images being investigated manually. To facilitate the classification process, all landmark images are studied carefully to select the highest detectable features with Gabor filtering to be the candidate features. For example, the Eiffel tower images contain two regions which are assumed to have accurately detectable features. The image sets of these regions are fed into the neural network as the feature image set. For the Eiffel tower landmark, the network output neurons will set to two. These outputs appear in a mutually exclusive basis, i.e. if the features exist in an image, one of the outputs will be greater than 0.8 or less than -0.5 and vice versa.

The image classification process described in algorithm 1 is shown in in figure 7. Careful examination of the image set of the Eiffel tower landmark, for example, reveals two features that are nominated as candidate features that lead to a reasonable extraction and hence; a good classification. The training image sets of these two areas are fed into the neural network as training sets. Figure 5 shows a sample image of the Eiffel tower, the two candidate feature regions, and the training sets fed into the network. The same process is done on the other four landmarks to select the dominant features and to prepare the training image sets of these features for next steps.

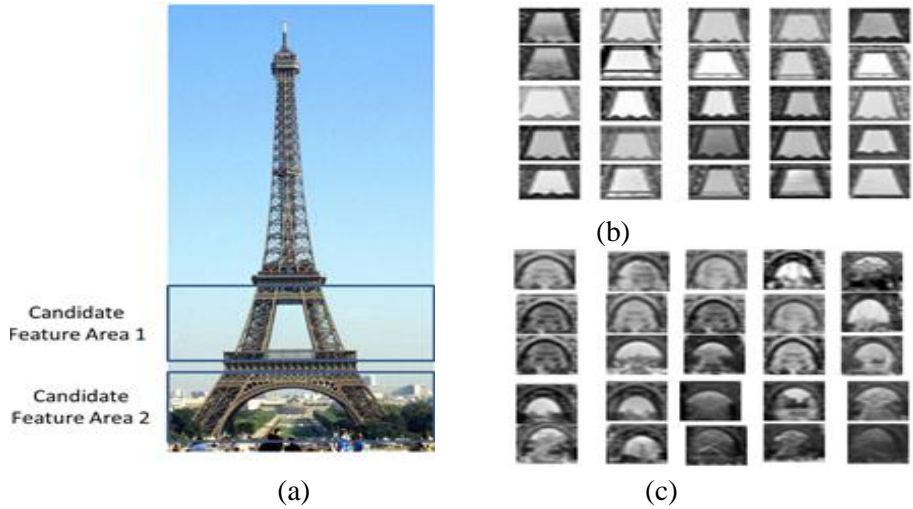

Figure 5. Eiffel tower image (a) Two candidate feature areas   (b) Sample of training set for area 1 (c) Sample of training set for area 2

### 5.4. Performing Gabor filtering

The Gabor filtering is performed by convolving images with Gabor kernels. Generally, convolution can be done using Fast Fourier Transform (FFT). The FFT is done by multiplying the frequency domain of Gabor kernels with the image pixel by pixel, i.e. dot product. Then, the inverse Fourier Transform (IFFT) has to be done to return the result back into spatial domain. Finally, the feature vector is built by converting the image data which was convolved with the Gabor kernels to prepare these images to be trained on the neural network. Figure 6 summarizes the feature extraction process.

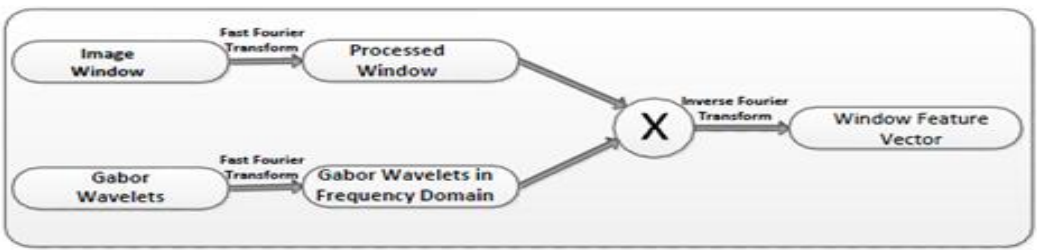

Figure 6. Feature Extraction Process

### 5.5. Candidate feature selection

For every landmark, the candidate feature(s) should be selected prior to training them on the neural network. The selection process depends on the nature of the landmark image. The candidate feature should be unique in the image, i.e. repeated features should not be selected. Also, the features with a larger texture similarity serve better in the selection process. Currently, this is a manual process.

## 5.6. Image matching process

The image set of a landmark downloaded or collected from the internet is pre-processed by resizing them and adjusting the contrast, brightness and illumination. Then, these images are applied to the neural network recognition phase to classify them as matched images or not. The images are placed in the matched folder or un-matched folder according to algorithm 1 in figure 7.

*Algorithm 1:*
 *Read the test feature-images.*
 *Read the test non-feature images.*
 *Set the number of candidate features.*
 *Set the training parameters:*
   *Epoch=300;*
   *Performance: MSE=0.3 x 10-3;*
   *Gradient= 10-6;*
 *Set neuron number to 1,*
 *Set all Neuron Detection factors to zero.*
 *Set Overall Matching factor to 1.*
 *While (neuron number ≤ number of candidate features)*
 *{*
     *Read neuron output;*
     *If (neuron output ≥ 0.8 )*
        *Set Neuron Detection factor to 1.*
     *Multiply Overall Matching factor by   Neuron Detection factor.*
     *Increment neuron number.*
 *}*
 *If (Overall Matching factor =1)*
  *Classify image as Matched images;*
 *Else*
  *Classify image as Un-Matched image*

Figure 7. Image classification algorithm

## 5.7. Experimental Results

The algorithm was applied to two different sets of images of five landmarks, Eiffel Tower, Coliseum of Rome, Pyramids of Giza, Dome Rock, and the Statue of Liberty. The first bunch of image data is the primary data set which comprised of 300-350 images for each of the landmarks. These images are preprocessed to improve their contrast and lighting and then cropped to maintain the most relevant parts.

The features that are most likely detected along with an example of problematic false positives are shown in figure 8. The false positive problem was partially addressed by selecting the features with highest matching scores. This leads to the selection of most unique features. The landmarks image sets were selected randomly from image websites as a mix of modern and historic images of five landmarks as well as non-landmark images. The landmark images had selected one, two, or three candidate features for evaluation. Each landmark set was applied to the system to classify them into proper folders, namely the landmark image folder and the non-landmark image folder.

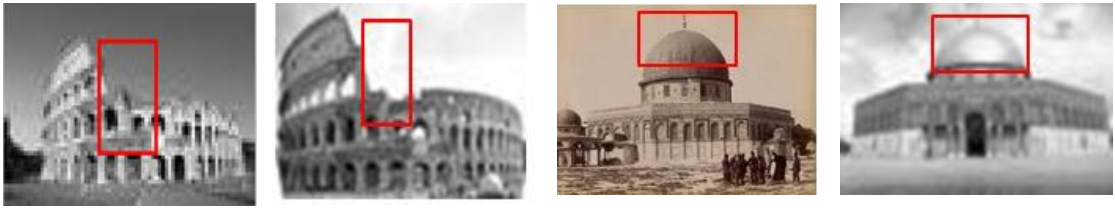

(a)

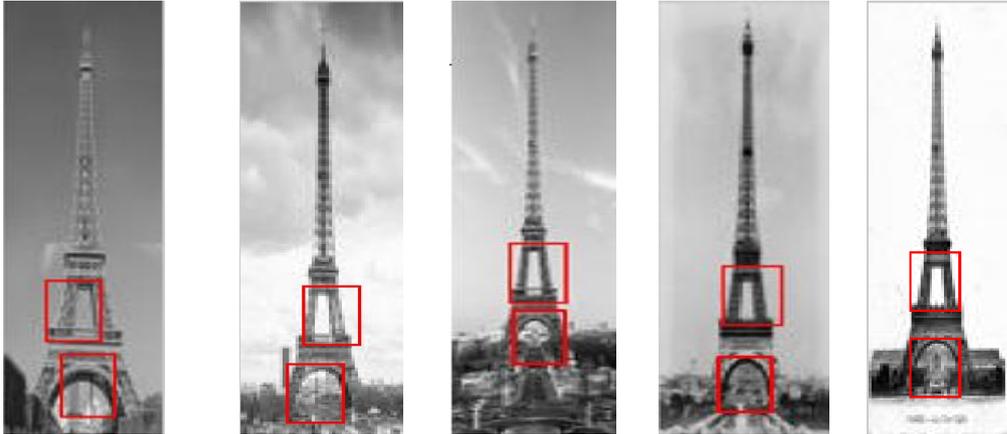

(b)

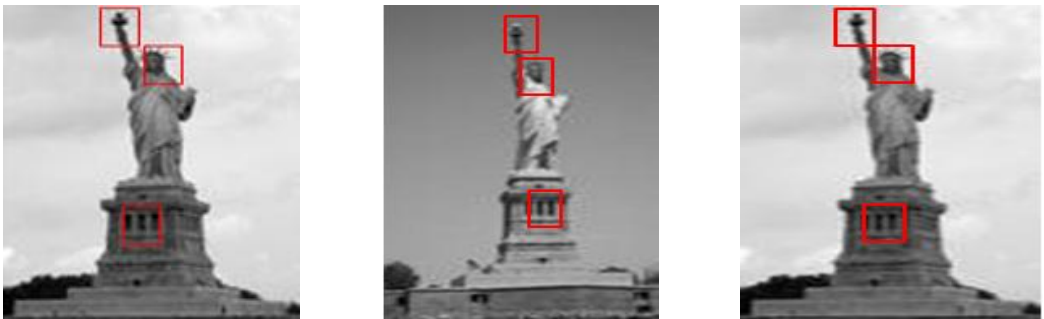

(c)

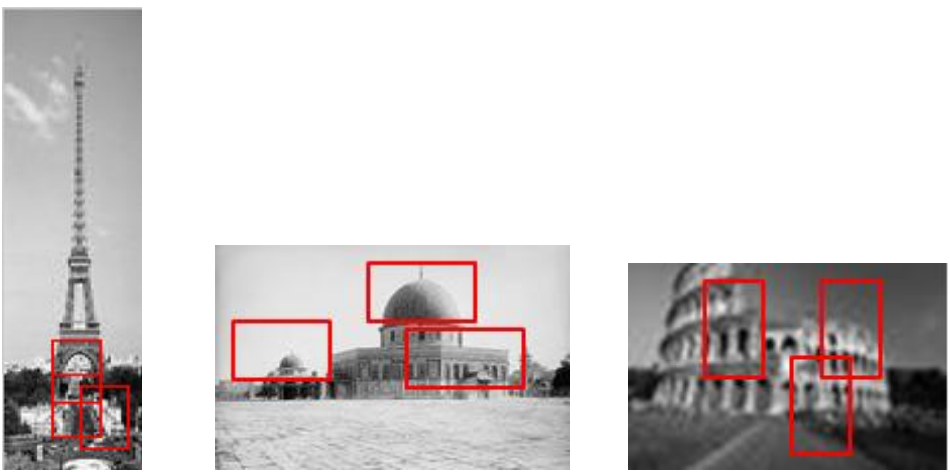

(d)

Figure 8. Feature extraction of images (a) 1- Feature (b) 2-Features (c) 3-Features (d) False positive features

The one-feature landmark detection results are shown in table 1 below:

Table 1. Applied algorithms results on 1-feature landmarks images

| Landmark | Total Images | True Positive | False Negative | False Positive | True Negative |
|---|---|---|---|---|---|
| Coliseum | 1520 | 532 | 301 | 412 | 188 |
| Dome | 1313 | 491 | 333 | 407 | 169 |

The two-feature landmark results are shown in table 2:

Table 2. Applied algorithms results on 2-feature landmarks images

| Landmark | Total Images | True Positive | False Negative | False Positive | True Negative |
|---|---|---|---|---|---|
| Eiffel | 1630 | 805 | 325 | 236 | 129 |
| Pyramid | 1330 | 612 | 455 | 285 | 113 |

While table 3 shows applying the algorithm on three-feature landmark images:

Table 3. Applied algorithms results on 3-feature landmarks images

| Landmark | Total Images | True Positive | False Negative | False Positive | True Negative |
|---|---|---|---|---|---|
| Statue | 2219 | 1569 | 331 | 216 | 103 |

A review of tables 1, 2 and 3 shows that the landmark image contents affect both the number of selected features and the image classification process, i.e. if the image objects seem as sparse or condensed highly affect the selection process. For example, in the Eiffel tower landmark, the main object in the images is the tower which is a single rigid object with parts could be extracted as features easily. Also, the non-object area of the images of Eiffel tower represents a background that facilitates the selection process as shown in figure 9-a. While images of pyramid landmark show objects that are not unique in nature and seem to be condensed with a repeated shape objects and this leads to complicate the classification process.

The main landmark object of Eiffel tower surrounded by a red line in figure 9-a represents an object that easy extracted with unique parts as shown in figure 5-a, while the Pyramid landmark shown in figure 9-b shows overlapped and interfered objects of blue line that are difficult to be extracted and with low overall accuracy.

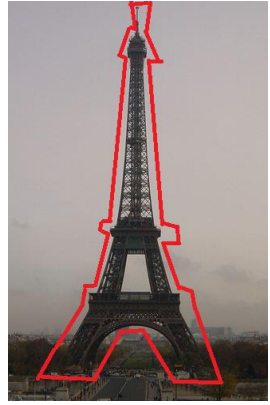 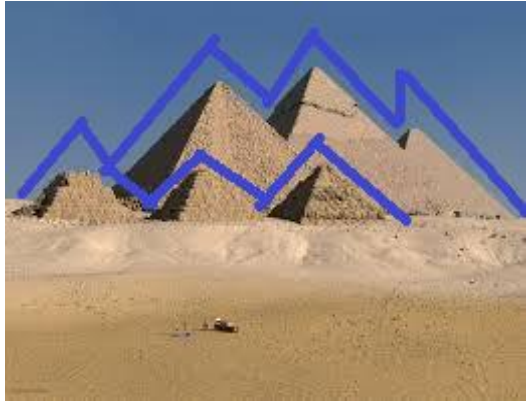

(a)                                 (b)

Figure 9. Object characteristics of (a) Eiffel tower and (b) Pyramid landmarks

The overall results of applying the algorithm on all landmarks data sets shown in table 4.

Table 4. Applied algorithms results on landmark images collected from the image websites

|  | Total Images | True Positive | False Negative | False Positive | True Negative |
|---|---|---|---|---|---|
| One Feature Image Set | 2833 | 1023 | 634 | 819 | 357 |
| Two Features Image Set | 2960 | 1417 | 780 | 521 | 242 |
| Three Features Image Set | 2219 | 1569 | 331 | 216 | 103 |

The results shown in table 4 declare that this algorithm performed effectively on a mixed of modern and historic images of a large number of five landmarks collected randomly from the Google images. Despite the diverse capturing environments for these images such as camera technology, lighting conditions, and geographical variations, this algorithm exploits the merging of Gabor filtering with the neural network to extract the dominant image features and to categorize images as belonging to a specific landmark image set or not.

To analyze the obtained results of tables 1,2,3 and 4, the precision and recall metrics were applied to calculate the precision, recall, accuracy and the F-measure factor which is the harmonic mean of precision and recall [27] using the formulas given in tables 5.

Table 5. Precision and recall metrics formulas

| Precision | Recall | Accuracy | F1 Score |
|---|---|---|---|
| tp/(tp+fp) | tp/(tp+fn) | (tp+tn)/(tp+tn+fp+fn) | 2*tp/(2*tp+fp+fn) |

## 5.8. Subset Selection Mechanism Performance and Effectiveness

The results shown in table 6,7,8 and 9 reveals that the precision and the accuracy of this mechanism increased as the number of the candidate features of landmark images increased, and the overall performance enhanced by increasing the number of the extracted features.

The selection mechanism applied on five sets of landmarks images picked up the images with highly detectable features despite the considerable variances in images views, scales, and lighting conditions in an automated manner shows that this subset selection mechanism effectiveness is reasonably accepted and represents a promising image set categorization and selection.

Table 6. Accuracy table of the algorithm of 1-feature landmarks image sets

|  | Precision | Recall | Accuracy | F1 Score |
|---|---|---|---|---|
| **Coliseum** | 0.563559322 | 0.638655462 | 0.502442428 | 0.598761958 |
| **Dome** | 0.546770601 | 0.595873786 | 0.471428571 | 0.570267131 |

Table 7. Accuracy table of the algorithm of 2-feature landmarks image sets

|  | Precision | Recall | Accuracy | F1 Score |
|---|---|---|---|---|
| **Eiffel** | 0.773294909 | 0.712389381 | 0.624749164 | 0.741593736 |
| **Pyramid** | 0.682274247 | 0.573570759 | 0.494880546 | 0.623217923 |

Table 8. Accuracy table of the algorithm of 3-feature landmarks image sets

|  | Precision | Recall | Accuracy | F1 Score |
|---|---|---|---|---|
| **Statue** | **0.878991597** | **0.825789474** | **0.753492564** | **1.482986767** |

Table 9. Accuracy table of the algorithm of all landmarks image sets

|  | Precision | Recall | Accuracy | F1 Score |
|---|---|---|---|---|
| **One Feature** | 0.555374593 | 0.617380809 | 0.487116131 | 0.584738497 |
| **Two Features** | 0.731166151 | 0.644970414 | 0.560472973 | 1.04267844 |
| **Three Features** | **0.878991597** | **0.825789474** | **0.753492564** | **1.482986767** |

## 6. CONCLUSIONS

Selection of the suitable images from a larger set of a landmark images containing both modern and historic images and captured with drastically different cameras under different lighting conditions and from different viewpoints represents a very important task when looking to match historical imaginary to modern imaginary. We show that Gabor filtering for such a task is a promising methodology when combined with a Neural Network for categorization. Dominant features from the images of many landmarks were selected as the training image set to be fed to the neural network. In the categorization stage, the raw images were applied as input to the neural network. The categorization results have shown the proposed method to be a promising first step in creating historical to modern timeline image sets [28].

## 7. FUTURE WORK

The overall performance of the algorithm can be enhanced by working on selecting more features from the input sets of landmark images by increasing the number of dominant features to four or even five since the overall accuracy of the algorithm increases as the number of the dominant increases. Also, a segmentation criterion may be applied on the input images to keep the most effective feature areas and/or to work on the overlapping objects separately.

# APPENDIX 1

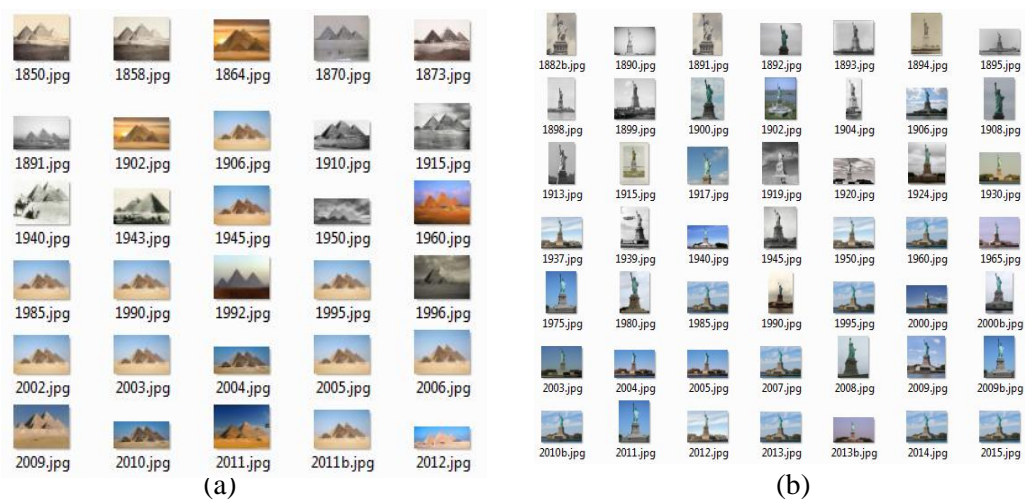

(a)     (b)

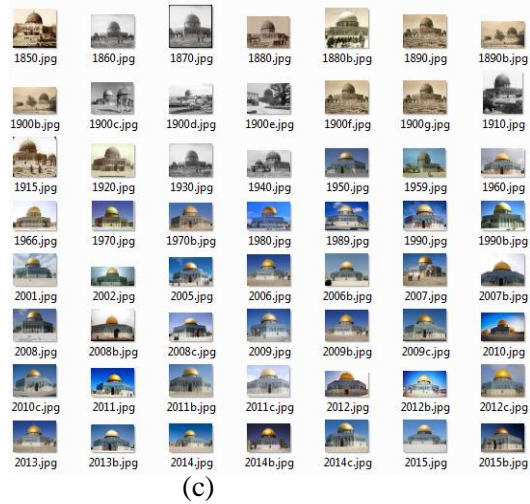

(c)

Figure A-1. Examples of modern and historic images of many landmarks (a) Pyramids of Giza (b) Statue of Liberty (c) Dome of Rock

## Authors

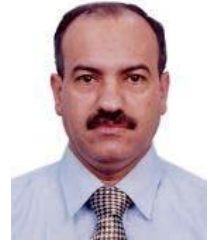

**Heider K Ali** is doing his PhD in Electrical and Computer Engineering at Carleton University. He works on the feature matching of modern to historic images of many worldwide landmarks and the investigation of the problems of dealing with historic photos. He got his MSc in computer engineering in 1991 and BSc in control and systems engineering in 1986 both from the University of Technology, Iraq. His fields of interest are computer vision, panorama building, historic to modern image matching, software engineering and machine learning.

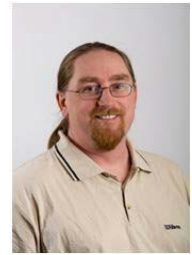

**Anthony Whitehead** is an Associate Professor and Director of the School of Information Technology at Carleton University. He is also the current Chair of the Human Computer Interaction Program at Carleton University. His research interests can be described as practical applications of machine learning, pattern matching and computer vision as they relate to interactive applications (including games), computer graphics, vision systems and information management.